\documentclass[10pt,twocolumn,letterpaper]{article}

\usepackage{iccv}
\usepackage{times}
\usepackage{epsfig}
\usepackage{graphicx}
\usepackage{amsmath}
\usepackage{amssymb}
\usepackage{booktabs}

\usepackage{xcolor}
\usepackage{color}
\usepackage{multirow}

\definecolor{blue}{rgb}{0.1,0.1,0.8}
\definecolor{orange}{rgb}{1,0.3,0.1}
\definecolor{yellow}{rgb}{1,1,0.8}

\usepackage[pagebackref=true,breaklinks=true,letterpaper=true,colorlinks,bookmarks=false,linkcolor=blue]{hyperref}

\iccvfinalcopy 
\def\iccvPaperID{2743} 

\begin{document}

\title{4D-Net for Learned Multi-Modal Alignment}

\author{AJ Piergiovanni\\
Google Research\\
\and
Vincent Casser\\
Waymo LLC\\
\and
Michael S. Ryoo\\
Robotics at Google\\
\and
Anelia Angelova \\
Google Research\\
}

\maketitle
\ificcvfinal\thispagestyle{empty}\fi

\begin{abstract}
  
We present 4D-Net, a 3D object detection approach, which utilizes 3D Point Cloud and RGB sensing information, both in time. We are able to incorporate the 4D information by performing a novel dynamic connection learning across various feature representations and levels of abstraction, as well as by observing geometric constraints.
Our approach outperforms the state-of-the-art and strong baselines on the Waymo Open Dataset.
4D-Net is better able to use motion cues and dense image information to detect distant objects more successfully.
We will open source the code.


\end{abstract}

\section{Introduction}

Scene understanding is a long-standing research topic in computer vision. It is especially important to the autonomous driving domain, where a central point of interest is detecting pedestrians, vehicles, obstacles and potential hazards in the environment. While it was traditionally undertaken from a still 2D image, 3D sensing is widely available, and most modern vehicle platforms are equipped with both 3D LiDAR sensors and multiple cameras producing 3D Point Clouds (PC) and RGB frames. Furthermore, autonomous vehicles obtain this information {\it in time}.
Since all sensors are grounded spatially, their data collectively, when looked at in time, can be seen as a 4-dimensional entity. 
Reasoning across these sensors and time clearly offers opportunities to obtain a more accurate and holistic understanding, instead of the traditional scene understanding from a single 2D still-image or a single 3D Point Cloud. 



\begin{figure} [t]
\vspace{-0.1cm}
    \includegraphics[width=1.0\linewidth]{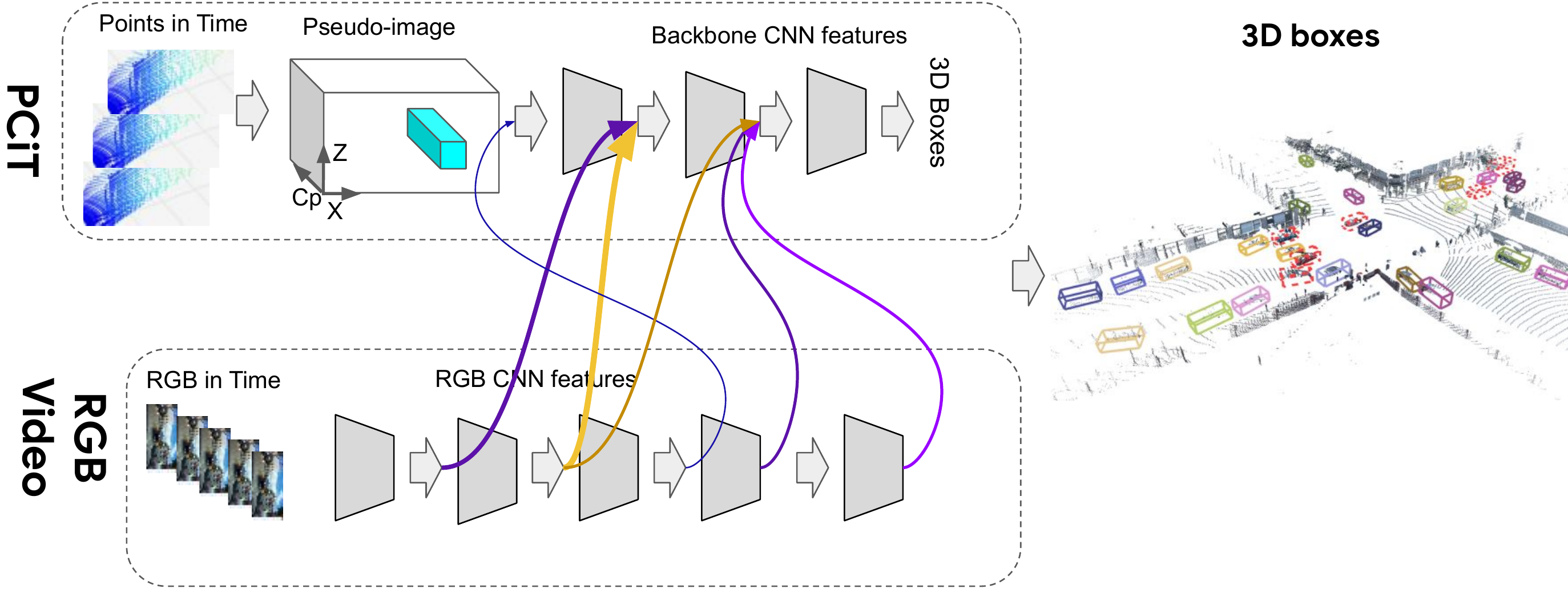}
    \caption{4D-Net effectively combines 3D sensing in time (PCiT) with RGB data also streamed in time, learning the connections between different sensors and their feature representations.}
     \label{fig:main}
\end{figure}

While all this 4D sensor data is readily available onboard, 
very few approaches have utilized it. For example, the majority of methods targeting 3D object detection use {\it a single} 3D point cloud as an input~\cite{Geiger2013IJRR}, with numerous approaches proposed~\cite{lang2019pointpillars,wang2020pillarbased,qi2017pointnet,ngiam2019starnet,shi2020pvrcnn,shi20193d,yan2018second,yan2018pixor,yang2018hdnet}. 
Only more recently has point cloud information been considered in time, with approaches typically accumulating several point clouds over a short time horizon~\cite{huang2020lstm,hu2020you,yan2018second,ngiam2019starnet}. 

Furthermore, the sensors have complementary characteristics. The point cloud data alone may sometimes be insufficient, e.g., at far ranges where an object only reflects a handful of points, or for very small objects. More information is undoubtedly contained in the RGB data, especially when combined with the 3D Point Cloud inputs. Yet, relatively few works attempted to combine these modalities~\cite{premebida2014pedestrian,vora2020pointpainting,gupta2014learning,ku2018joint}. Notably, only 2 of the 26 submissions to the Waymo Open Dataset 3D detection challenge operated on both modalities~\cite{sun2020scalability}. No methods have attempted combining them when both are streamed in time.
The questions of how to align these very different sensor modalities most effectively, as well as how to do so efficiently, have been major roadblocks.

\begin{figure*}
    \centering
    \includegraphics[width=0.9\linewidth]{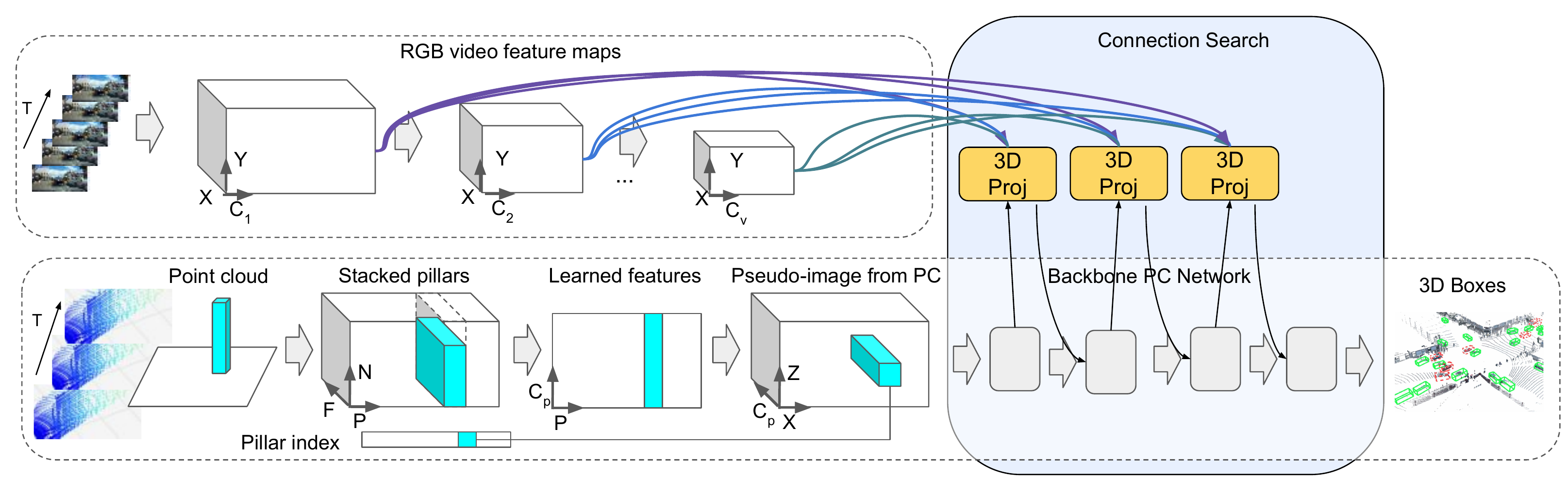}
    \caption{4D-Net Overview. RGB frames and Point Clouds in time are processed producing features, abstracting some dimensions. A connection search learns where and how to fuse the modalities' features together. We use 3D projection to align the PC and RGB features. 
    }
    \label{fig:overview}
\end{figure*}

To address these challenges, we propose {\it 4D-Net}, which combines Point Cloud information together with RGB camera data, both in time, in an efficient and learnable manner. We propose a novel learning technique for fusing information in 4D from both sensors, respectively building and learning connections between feature representations from different modalities and levels of abstraction (\autoref{fig:main}). 
Using our method, each modality is processed with a suitable architecture producing rich features, 
which are then aligned and fused at different levels by dynamic connection learning (\autoref{fig:overview}).
We show that this is an effective and efficient way of processing 4D information from multiple sensors.
4D-Nets provide unique opportunities as they naturally learn to establish relations between these sensors' features, combining information at various learning stages. This is in contrast to previous late fusion work, which fuse already mature features that may have lost spatial information, crucial to detecting objects in 3D. 

Our results are evaluated on the Waymo Open Dataset~\cite{sun2019scalability}, a challenging Autonomous Driving dataset and popular 3D detection benchmark. 4D-Net outperforms the state-of-the-art 
and is competitive in runtime. Importantly, being able to incorporate dense spatial information and information in time improves detection at far ranges and for small and hard to see objects. 
We present several insights into the respective significance of the different sensors and time horizons, and runtime/accuracy trade-offs.

Our contributions are: \textbf{(1)} the first 4D-Net for object detection which spans the 4-Dimensions, incorporating both point clouds and images in time, \textbf{(2)} a novel learning method which learns to fuse multiple modalities in 4D, \textbf{(3)} a simple and effective sampling technique for 3D Point Clouds in time, 
\textbf{(4)} a new state-of-the-art for 3D detection on the Waymo Open Dataset and a detailed analysis for unlocking further performance gains. 

\section{Related Work}





\textbf{Object Detection from RGB.}\mbox{} 
The earliest detection approaches in the context of autonomous driving were primarily focused on camera-based object detection, often drawing heavily from the extensive body of 2D vision work~\cite{viola2003detecting,dollar2009pedestrian,benenson2014ten,dollar2009pedestrian,benenson2012pedestrian, brazil2017illuminating,ouyang2012a,cai2016a,stewart2016e2e,Geiger2013IJRR,Cordts2016Cityscapes}, with some more advanced works using deep learning features~\cite{angelova2015realtime,chu2020detection,song2020progressive}.  
Detection with temporal features, i.e., integrating features across several neighboring frames~\cite{enzweiler2011a,wu2020temporal}, and leveraging kinematic motion to improve detection consistency across time~\cite{brazil2020kinematic} have also been applied.
However, looking at the images as videos and processing them with video CNNs is not common. 


\textbf{Object Detection from Point Cloud.}\mbox{} 
Many approaches found it effective to apply well-established 2D detectors on a top-down (BEV) projection of the point cloud (AVOD~\cite{ku2018joint}, PIXOR~\cite{yan2018pixor}, Complex-YOLO~\cite{simon2018ComplexYOLO}, HDNET~\cite{yang2018hdnet}). This input representation can be advantageous because it makes it easier to separate objects, and object sizes remain constant across different ranges. However, it results in loss of occlusion information, does not effectively exploit the full 3D geometric information, and is inherently sparse. 
Certain techniques can be used to alleviate these drawbacks, e.g., learning a pseudo-image projection~\cite{lang2019pointpillars}. 

An alternative to top-down representations is operating directly on the range image, projecting the PC into perspective view. This representation is inherently dense, simplifies occlusion reasoning and has been used in various works (e.g. LaserNet~\cite{meyer2019lasernet}, VeloFCN~\cite{li2016vehicle}, \cite{bewley2020range}). Another line of work tries to exploit the complementary nature of both by jointly operating on multiple views~\cite{zhou2019multiview,wang2020pillarbased}. 

Instead of relying on view projections, some methods directly operate on the 3D voxelized point cloud~\cite{zhou2018voxelnet,li20173d,wang2015voting}. Voxel resolution can greatly affect performance and is typically limited by computational constraints. To reduce compute, some rely on applying sparse 3D convolutions, such as Vote3Deep~\cite{engelcke2017vote3deep}, Second~\cite{yan2018second} or PVRCNN~\cite{shi2020pvrcnn}. Dynamic voxelization has been proposed in~\cite{zhou2019multiview}.

Other methods operate directly on the raw point cloud data, e.g. SPLATNet~\cite{su2018SplatNet}, StarNet~\cite{ngiam2019starnet} or PointRCNN~\cite{shi20193d}.

\textbf{Point Clouds in Time (PCiT).}\mbox{} 
Integrating information from multiple point clouds has been proposed recently. StarNet \cite{ngiam2019starnet} does not explicitly operate in time, but can use high-confidence predictions on previous frames as ``temporal context'' to seed object center sampling in the following frames. \cite{huang2020lstm} extract features on individual frames and accumulate information in an LSTM over 4 frames for detection. 
\cite{choy20194d} apply 4D ConvNets for spatial-temporal reasoning for AR/VR applications.
\cite{yan2018second,hu2020you,zhang2020stinet} combine multiple point clouds in time by concatenating them and adding a channel representing their relative 
timestamps.

\textbf{Point Clouds and RGB fusion.}\mbox{} 
Acknowledging the merits of sensor fusion, researchers have attempted to combine LiDAR and camera sensing to improve performance~\cite{premebida2014pedestrian,chen2017multi,liang2019multiview}. 
Frustum PointNet~\cite{qi2018frustum} first performs image-based 2D detection, and then extrapolates the detection into a 3D frustum based on LiDAR data.
Alternatively, one can project the point cloud into the camera view - in its simplest form creating RGB-D input, although alternative depth representations may be used~\cite{gupta2014learning,girshick2014rich}. However, this provides limited scalability as each detection inference would only cover a very limited field of view. Conversely, the point cloud input can be enriched by adding color or semantic features~\cite{vora2020pointpainting,song2014sliding,ku2018joint}. This, however, comes at the expense of losing spatial density - one of the primary advantages of camera sensors.
%
%
%
%
%
Several methods have instead applied modality-specific feature extractors, which are then fused downstream~\cite{ku2018joint,liang2018deep,chen2017multi,xu2018pointfusion,liang2019multiview}. 
None of the abovementioned approaches operate on either modality in time.




\section{4D-Net}

4D-Net proposes an approach to utilize and fuse multi-sensor information, learning the feature representation from these sensors and their mutual combinations.
An overview of the approach is shown in \autoref{fig:overview}.
In 4D-Net, we consider the point clouds in time (i.e., a sequence of point clouds) and the RGB information, also in time (a sequence of images). 
We first describe how to handle the raw 3D point clouds and RGB input information streaming in time (Section~\ref{sec:processing}) and then describe our main 4D-Net architectures which learn to combine information from both sensor modalities and across dimensions (Section~\ref{sec:4d}).
We further offer a multi-stream variant of the 4D-Net which achieves further performance improvements 
(Section~\ref{sec:multistream}).



    
     
     


\subsection{3D Processing and Processing Data in Time}
\label{sec:processing}
\vspace{-0.1cm}
\subsubsection{3D Processing}
\label{sec:pointpillars}
\vspace{-0.1cm}
Our approach uses a learnable pre-processor for the point cloud data; it is applied to the 3D points and their features from the LiDAR response to create output features. We chose to use PointPillars \cite{lang2019pointpillars} to generate these features, but other 3D point `featurising' approaches can be used. PointPillars converts a point cloud into a pseudo-image, which can then be processed by a standard 2D CNN. 
For clarity, in the derivations below, we will be using a 3D $\mathcal{X}, \mathcal{Y}, \mathcal{Z}$, coordinate system, where the $\mathcal{Z}$ direction is forward (aligned with the car driving), $\mathcal{Y}$ is vertical pointing up and $\mathcal{X}$ is horizontal, i.e., we use a left-hand coordinate system (this is the default system used in the Waymo Open Dataset). 

Given a point cloud $\mathcal{P} = \{p\}$ where $p$ is a 3D $(x,y,z)$ point and associated $F$-dimensional feature vector (e.g., intensity, elongation), the pseudo-image is created as follows. Each point is processed by a linear layer, batch norm and ReLU, to obtain a featurized set of 3D points. The points are grouped into a set of pillars in the $\mathcal{X}$, $\mathcal{Z}$ plane based on their 3D location and distances between the points. This gives a point cloud representation with shape $(P, N, F)$, where $P$ is the number of pillars, and $N$ is the maximum number of points per pillar. Each of the $P$ pillars is associated with a $x_0,y_0,z_0$ location that is the pillar center. 
The idea is then to further `featurize' information in this $(P, N, F)$ representation and then, using the original coordinates, to `distribute' back the features along the $\mathcal{X}$, $\mathcal{Z}$ plane and produce a pseudo-image~\cite{lang2019pointpillars}, say of size  $(X, Z, C_P)$. Specifically, from $(P, N, F)$, a feature of size $(P, C_P)$ is obtained via learnable layers and pooling, to then get $(X, Z, C_P)$.
In effect, PointPillar produces a $(X,Z,C_P)$ feature representation from a $(X,Y,Z,F)$ input for a single PC.
%
%

\vspace{-0.3cm}
\subsubsection{Point Clouds in Time}
\vspace{-0.1cm}
Point clouds and the subsequent feature creation (e.g., as in Section~\ref{sec:pointpillars}) are computationally expensive and memory intensive operations. Given a sequence of $T$ point clouds, creating $T$ PointPillar ``pseudo-images'' and then using a 2D or 3D CNN to process all those frames would be prohibitively expensive~\cite{zhang2020stinet}, limiting its usefulness. Previous work \cite{huang2020lstm} explored using sparse convolutions with LSTMs to handle point clouds in time, where a compressed feature representation is fed recursively to next frame representations. Instead, we take a simpler approach similar to~\cite{caesar2020nuscenes}, which however preserves the original feature representation per 3D point, together with a sense of time.

First, the original feature representation is directly merged in the 3D point cloud, together with a feature to indicate its timestamp.
Specifically, we use the vehicle pose to remove the effect of ego-motion and align the point clouds. Next, we add a time indicator $t$ to the feature of each point: $p = [x, y, z, t]$. Then, the PointPillar pseudo-image representation is created as before, which also results in a denser representation. While dynamic motion will obviously create a ghost/halo effect, it can in fact be a very useful signal for learning, and be resolved by the time information. In some circumstances, it is only through motion that distant or poorly discernible objects can be detected. 

\begin{figure}
    \centering
    \includegraphics[width=\linewidth]{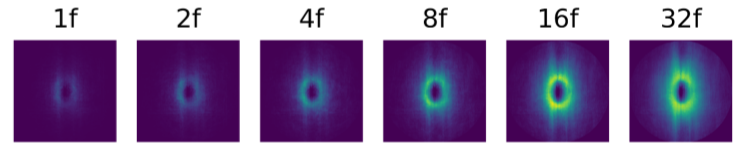}
    \caption{
    The average point density per voxel illustrates how long-term temporal aggregation, combined with our subsampling strategy, leads to an increased point density, especially at far ranges.
    }
    \label{fig:pc-density}
\vspace{-0.4cm}
\end{figure}

\textbf{Point Cloud Subsampling.} Importantly,
when accumulating points from multiple point clouds, the voxelization step converts all the points into a {\it fixed-sized} representation based on the grid 
cell size. This results in a tensor with a fixed size that is padded to $N$, the maximum number of points. Thus the amount of subsequent compute remains the same, regardless of the number of PCs. If the points exceed $N$, only $N$ points are randomly sampled ($N$=128 throughout). 
This has the effect of subsampling the accumulated point cloud, but proportionally more points will be sampled in sparser areas than in dense ones. By densifying the point cloud in sparse regions and sparsifying it in dense regions, we distribute compute more efficiently and provide more signal
for long-range detection  by increased {\it point density at  far ranges} (e.g. see point density for 16 PCs in \autoref{fig:pc-density}). 
%
We find this representation to be very effective, resulting in significant improvements over using a single point cloud (as seen later in ablation \autoref{tab:ablation-pc}).

\begin{figure}[t]
    \centering
    \includegraphics[width=\linewidth]{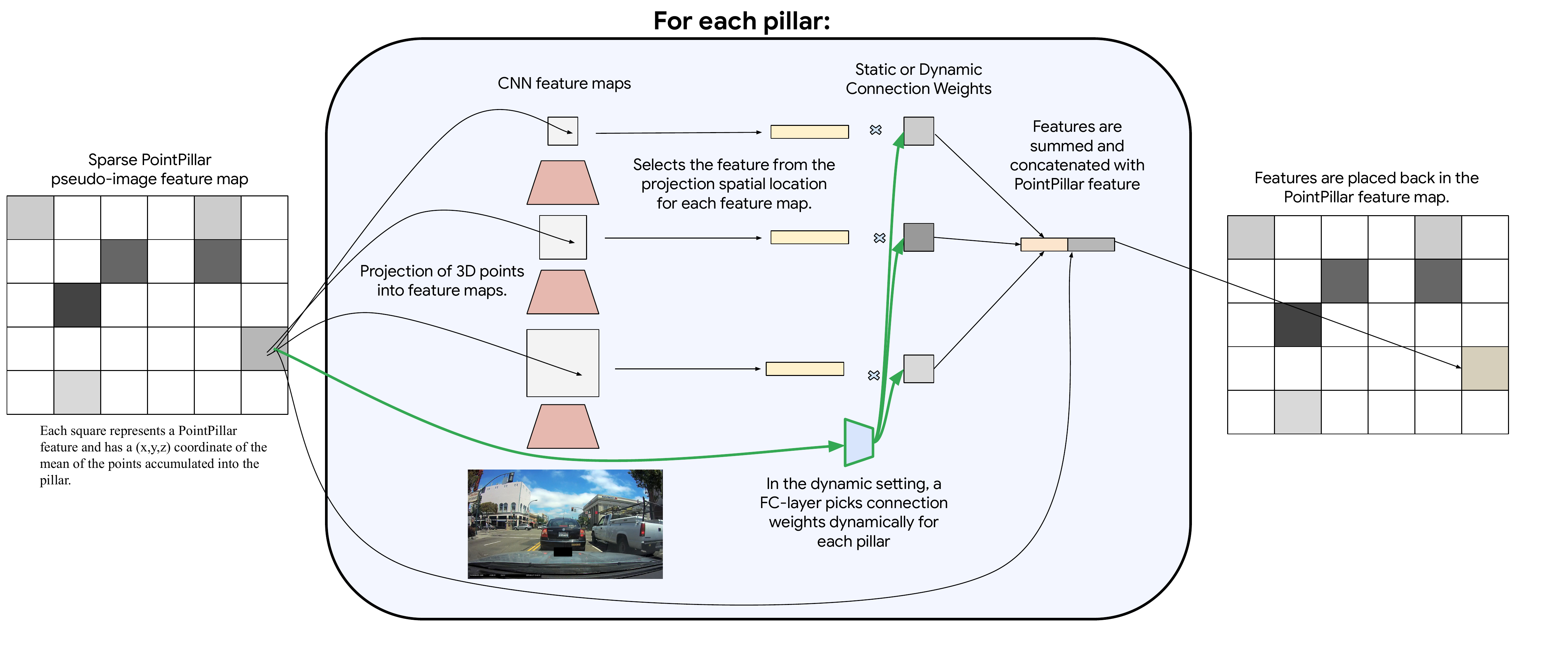}
    \caption{
    \textbf{Connection architecture search}: Each PointPillar feature is projected into the CNN feature map space based on the 3D coordinate of each pillar and the given camera matrices. The feature from the 2D location is extracted from each feature map. Learned static connection weights (grey boxes) then combine these and concatenate them with the pillar feature to create the pillar feature map used as input to the next layer. \textbf{Dynamic connections}, shown in green, are produced for each value which are controlled by the features from the point clouds, thus they determine how to fuse in the rest of the features generated by the model at various levels of abstractions. Please see 
    Section~\ref{sec:arch_search}.
    }
    \label{fig:dynamic-connections}
\vspace{-0.4cm}
\end{figure}


\vspace{-0.4cm}
\subsubsection{RGB and RGB in Time}
\vspace{-0.2cm}
While point clouds have become the predominant input modality for 3D tasks, RGB information is very valuable, especially at larger distances (e.g., 30+ or 50+ meters) where objects garner fewer points.

Furthermore, images in time are also highly informative, and complementary to both a still image and PCiT. In fact, for challenging detection cases, motion can be a very powerful clue. While motion can be captured in 3D, a purely PC-based method might miss such signals simply because of the sensing sparsity.

RGB frames, unlike PCs, represent a dense feature containing color pixel information for everything in view. Here we take RGB frames as input and use video CNNs to produce the RGB feature maps. In the video settings, we take $T$ previous frames as input and predict the object bounding boxes in the final frame, same as for point clouds in time.
We process a sequence of RGB frames as a video input. Since runtime is of the essence, we use efficient video representations, Tiny Video Networks \cite{piergiovanni2020tvn} for processing. More specifically, by a series of layers, some of which working in the spatial dimensions, some temporal, a set of feature representations in the spatial dimension will be learned.
As a result, the feature with shape ($X,Y,C_V$) is produced, abstracting away the time coordinate in input ($X,Y,T,3$).

\vspace{-0.1cm}
\subsection{4D-Net: Fusing RGB in Time and PC in Time}
\label{sec:4d}

To combine RGB information into the 3D Point Cloud PointPillar representation (both in time), there are two major considerations: 1) the two sensors need to be geometrically and spatially aligned and 2) the fusion mechanisms of the features produced from these modalities should ideally be learned from the data. 

Our 4D-Net entails both projection fusion mechanisms and connectivity search to learn where and how to fuse features (\autoref{fig:overview}). Of note is that both representations have abstracted away some `dimensions' in their features but still contain complementary information: the RGB representation has ($X$, $Y$, $C_V$), whereas the PCiT has ($X$, $Z$, $C_P$).
 Since our end goal is 3D object detection, we chose to fuse from RGB into the point cloud stream, but we note that these approaches could be used to fuse in the other direction as well.
 Section~\ref{sec:proj} and Section~\ref{sec:arch_search}, provide details.

\vspace{-0.3cm}
\subsubsection{3D Projection} 
\label{sec:proj}
To fuse the RGB into the point cloud, we need to (approximately) align the 3D points with 2D image points. 
To do this, we assume we have calibrated and synchronized sensors and can therefore define accurate projections. 
Note that the Waymo Open Dataset provides all calibration and synchronized LiDAR and camera data.

The PointPillar pseudo-image $M$ has shape $(X, Z, C_P)$ and is passed through a backbone network with a ResNet-like structure. After each residual block, $i$, the network feature map $M_i$ has shape $(X^M_i, Z^M_i, C^M_i)$, where each location (in $\mathcal{X}$ and $\mathcal{Z}$) corresponds to a pillar. Each pillar $p$ also has an $(x_0,y_0,z_0)$ coordinate representing its center based on the accumulated 3D points. This provides a 3D coordinate for each of the non-empty feature map locations.

The RGB network also uses a backbone to process the video input. Let us assume that after each block, the network produces a feature map $R_i$ with shape $(X^R_i, Y^R_i, C^R_i)$, which is a standard image CNN feature map.

Using projections, we can combine the RGB and point cloud data. Specifically, given a $4\times 4$ homogeneous extrinsic camera matrix $E$ (i.e., the camera location and orientation in the world) and a $4\times 4$ homogeneous intrinsic camera matrix $K$ ($E$ and $K$ are part of the dataset), we can project a 3D point $p=(x,y,z,1)$ to a 2D point as $q = K\cdot(E\cdot p)$. For each point pillar location, we obtain the 2D point $q$, which provides an RGB feature for that point as $R_i[q_x, q_y]$\footnote{We tried a spatial crop around the point, but found it to be computationally expensive. The CNN's receptive field also provides spatial context.}. This is concatenated to the pillar's feature, e.g.,
\begin{equation}
    M_i[p_x, p_y] = [M_i[p_x, p_y] | R_i[K\cdot(E\cdot p)]] p\in\mathcal{P}
\end{equation}

Note that LiDAR data typically covers a full 360 degree surround view, while individual cameras typically have a quite limited horizontal field of view. To account for this, we only obtain RGB features for points which are captured by one of the cameras. For points outside of the image view, we concatenate a vector of zeros. This approach is easily applied to settings with multiple RGB cameras covering different viewpoints -- the same RGB CNN is applied to each view, then the projection is done per-camera, and added together before concatenation. 

\vspace{-0.3cm}
\subsubsection{Connection Architecture Search in 4D}
\label{sec:arch_search}
While the above projection will align the two sensors geometrically, it is not immediately obvious what information should be extracted from each and how the sensor features interact for the main task of object detection.

To that end we propose to learn the connections and fusion of these via a light-weight differentiable one-shot architecture search. One-shot differentiable architecture search has been used for strengthening the learned features for image understanding~\cite{liu2019darts} and for video~\cite{ryoo2020assemblenet,ryoo2020assemblenetplus}.
In our case, we are utilizing it for relating information in 4D, i.e., in 3D and in time, and also connecting related features across different sensing modalities (the RGB-in-time and point cloud-in-time streams).
Of note is that we learn the combination of feature representations at various levels of abstraction for both sensors (\autoref{fig:overview}).

In \autoref{fig:dynamic-connections}, we illustrate how this architecture search works. Given a set of RGB feature maps, $\{R_i | i\in [0,1,\ldots, B]\}$ ($B$ being the total number of blocks/feature maps in the RGB network), we can compute the projection of each pillar into the 2D space and obtain a feature vector. This produces a set of feature vectors $\mathcal{F} = \{f_i | i\in [0, 1,\ldots, B]\}$. We then have a learned weight $w$, which is a $B$-dimensional vector. We apply softmax and then compute $\sum w\times \mathcal{F}$ to obtain the final feature vector. $w$ learns the connection weights, i.e. which RGB layer to fuse into the PointPillars layer. This is done after each block in the PointPillars network, allowing many connections to be learned (\autoref{fig:overview}).

\begin{figure} [t]
\vspace{-0.1cm}
\begin{center}
    \includegraphics[width=0.9\linewidth]{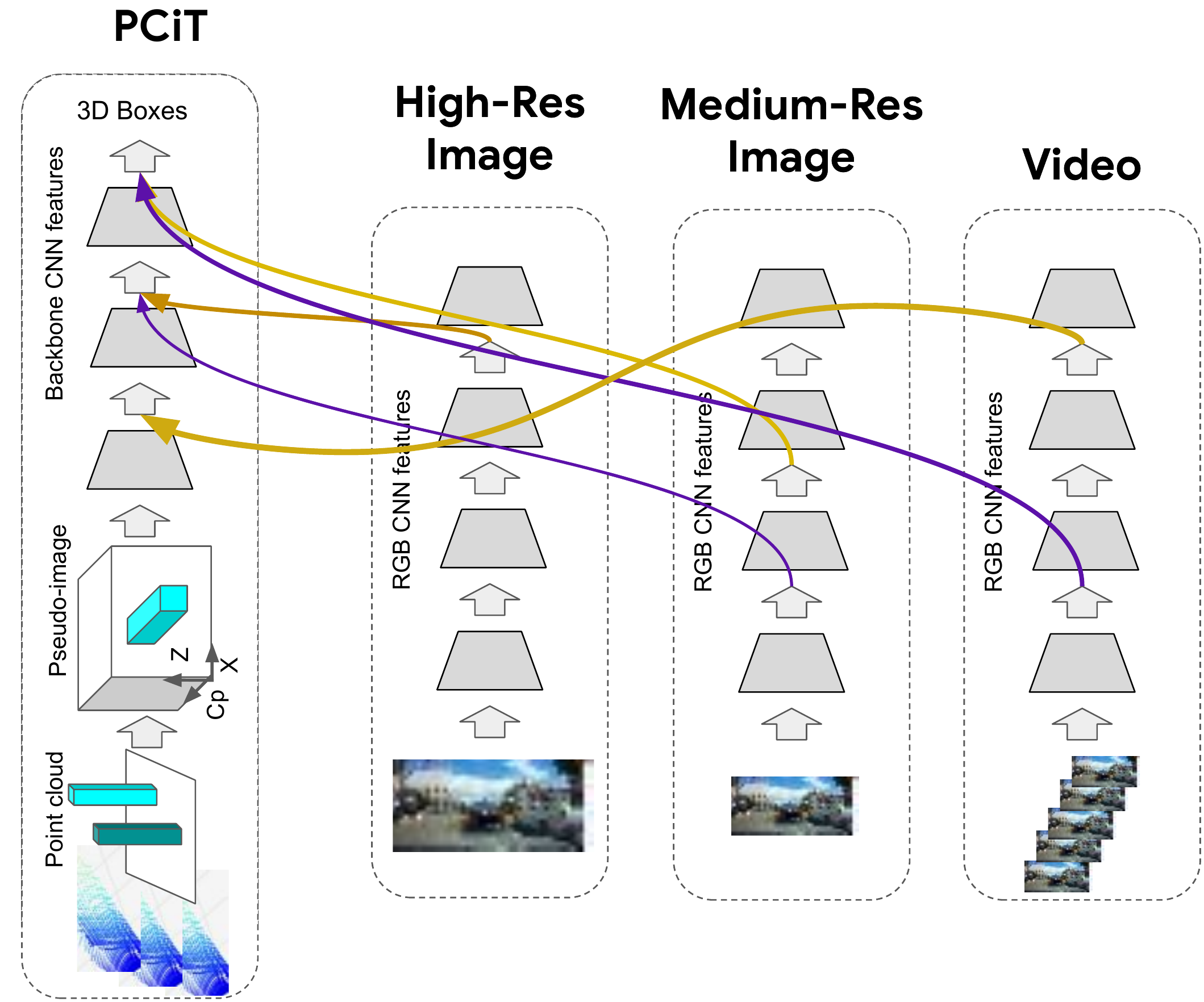}
    \end{center}
    \caption{Illustration of a multi-stream 4D-Net. It takes point clouds (in time) and still-image and video as input, computes features for them, learns connection weights between the streams.}
     \label{fig:multi}
\vspace{-0.25cm}
\end{figure}

\textbf{Dynamic Connections.}
The above-mentioned mechanism is very powerful as it allows to learn the relations between different levels of feature abstraction and different sources of features (e.g., RGB, PC) (\autoref{fig:dynamic-connections}). Furthermore, as shown in later sections (Section~\ref{sec:multistream}) it allows for combining multiple computational towers seamlessly without any additional changes.

However, in the autonomous driving domain it is especially important to reliably detect objects at highly variable distances, with modern LiDAR sensors reaching several hundreds of meters of range~\cite{waymo_new_sensor_announce}. This implies that further-away objects will appear smaller in the images and the most valuable features for detecting them will be in earlier layers, compared to close-by objects. Based on this observation, we modified the connections to be {\it \textbf{dynamic}}, inspired by self-attention mechanisms. Specifically, instead of $w$ being a learned weight, we replace $w$ with a linear layer with $B$ outputs, $\omega$, which is applied to the PointPillar feature $M_i[p_x, p_y]$ and {\it generates weights over the $B$ RGB feature maps}. $\omega$ is followed by a softmax activation function. This allows the network to dynamically select which RGB block to fuse information from, e.g. taking a higher resolution feature from an early layer or a low resolution feature from a later layer (\autoref{fig:dynamic-connections}). Since this is done for each pillar individually, the network can learn how and where to select these features based on the input.


\begin{table*}[]
    \centering
    \scalebox{0.932}{
     \begin{tabular}{l|cc|ccc|c}
    \toprule
    Method   & AP L1 & AP L2  & AP 30m & AP 30-50m & AP 50m+  & Runtime \\
    \midrule
    StarNet~\cite{ngiam2019starnet}   &53.7 & - & - & -  &- &- \\
    LaserNet~\cite{meyer2019lasernet}    & 52.1 & - & 70.9 & 52.9  &29.6  & 64ms\\
    PointPillars~\cite{lang2019pointpillars}, from~\cite{huang2020lstm} &57.2 &- &- &- &- &- \\
    MVF~\cite{zhou2019multiview} &62.9 &- &86.3 &60.0 &36.0 &- \\
    Huang et al~\cite{huang2020lstm} (4 PCs)   & 63.6 & - & -  &-  &- &-\\
    PillarMultiView~\cite{wang2020pillarbased}  & 69.8  & - & 88.5 & 66.5 & 42.9 &67ms \\
    PVRCNN~\cite{shi2020pvrcnn}    & 70.3 & 65.4 & 91.9 & 69.2 & 42.2 & 300ms  \\ 
    \midrule
    4D-Net (Ours)  & 73.6  &70.6	&80.7	&74.3	&56.8 & 142 ms (net) + 102 ms (16f voxel)\\
    4D-Net (Ours with Multi-Stream) &74.5   &71.2	&80.9	&74.7	&57.6 &  203 ms (net) + 102 ms (16f voxel)  \\
    \bottomrule
    \end{tabular}
    }
       \caption{Waymo Open Dataset~\cite{sun2020scalability}. 3D detection AP on vehicles @ 0.7 IoU on the validation set. For 4D-Net we report the runtime of the network and of the pre-processing voxelization step for the point clouds in time.}
    \label{tab:main}
\end{table*}

\begin{figure*} [t]
\vspace{-0.1cm}
    \includegraphics[width=1.0\linewidth]{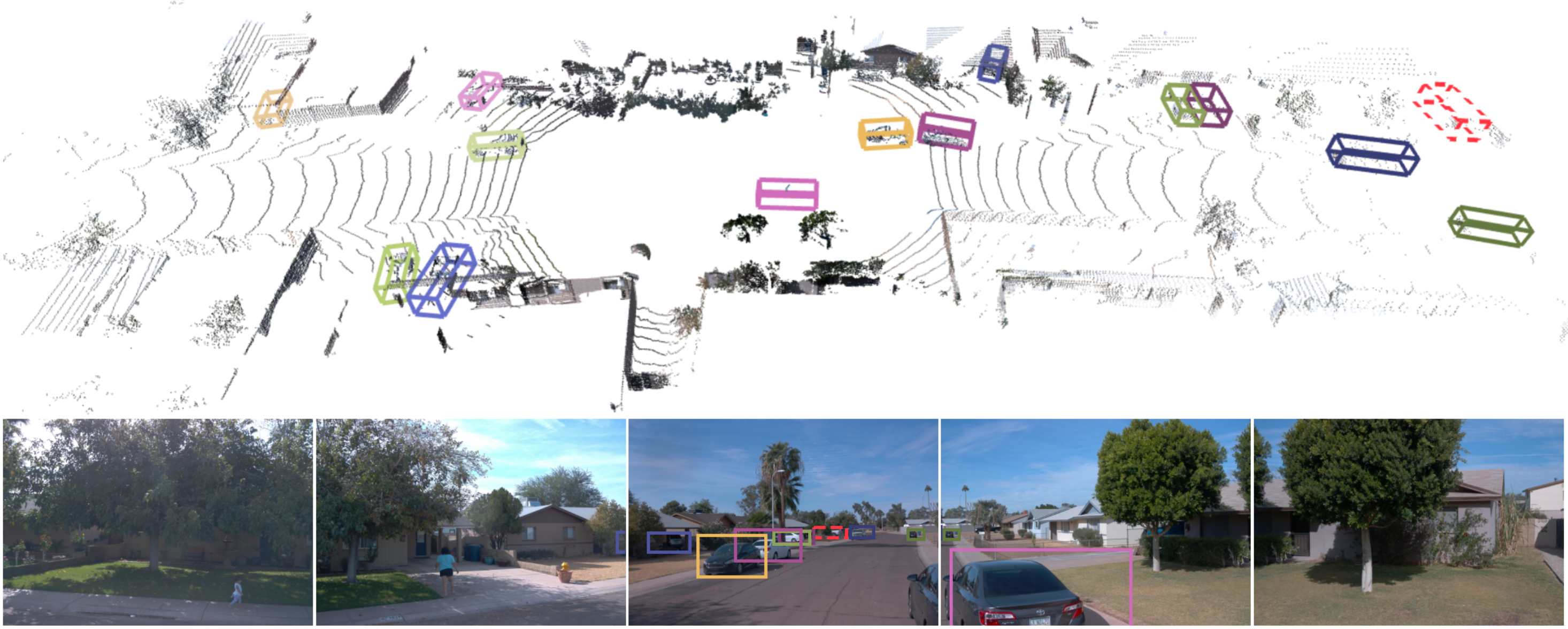}
    
    \caption{4D-Net predictions on a scene in the Waymo Open Dataset~\cite{sun2020scalability}. Individual instances are shown in different colors. Red boxes indicate errors (dashed lines: FN, solid lines: FP). In this example, all predictions are matching the ground-truth except for a false-negative on the right side (front camera). Note that any misalignments in the camera view are due to projection, not by inaccuracies in the predictions.}
     \label{fig:WOD}
\vspace{-3mm}
\end{figure*}

\subsection{Multi-Stream 4D-Net}
\label{sec:multistream}

\textbf{Multipe RGB streams.}
Building on the dynamic connection learning, we propose a Multi-Stream (MS) version of 4D-Net. While the 4D-Net itself already learns to combine the information from two streams -- the sparse 3D PCiT and camera input -- we can have more than one RGB input stream (\autoref{fig:multi}). One advantage of the proposed (dynamic) connection learning between features of different modalities is that it is applicable to many input feature sources and is agnostic to where they originate from. 
For example, we can add a separate tower for processing high-resolution still images, or an additional video tower using a different backbone or a different temporal resolution. This enables a more rich set of motion features to be learned and surfaced for combination with the PC features. 
Note that all these are combined with the PC (in time) features in the same dynamic fusion proposed above,  
thus allowing the PCiT stream to dynamically select the RGB features to fuse from all streams. Similarly, it is also possible to introduce additional PC streams.

\textbf{Multiple Resolutions.}
Empirically, we observe that adding an RGB stream benefits recognition of far away objects the most. Distant objects appear smaller than close objects, suggesting that using higher resolution images will further improve recognition. 
Additionally, adding RGB inputs at two or more different resolutions will increase the diversity of features available for connection learning.

Thus in the multi-stream setting, we combine inputs at different resolutions (see~\autoref{fig:multi} for a schematic). Our main Multi-Stream 4D-Net uses 
1) one single still image tower at high image resolution (312x312), 
2) one video image tower at lower resolution (192x192) with 16 frames
and 3) the PCiT which has aggregated 16 point clouds.
The original 4D-Net has the latter two streams only but uses 12 RGB frames.
More streams and more resolutions are explored in the ablations, Section~\ref{sec:ablation}.
%
%
Similar to our main 4D-Net, we use a lightweight Tiny Video Network~\cite{piergiovanni2020tvn}, so that the multi-stream 4D-Net is efficient at inference.

\vspace{-4mm}
\paragraph{Implementation Details.} We train the model to minimize a standard cross-entropy loss for classification and a $L_2$ regression loss for the residuals between the anchor boxes and the GT boxes. $N=128$ throughout the paper and we use a $224\times224\times1$ output grid.
Further implementation and experimental details are included in the sup. material.




\begin{figure*} [t]
\vspace{-0.1cm}

    

    \includegraphics[width=0.48\linewidth]{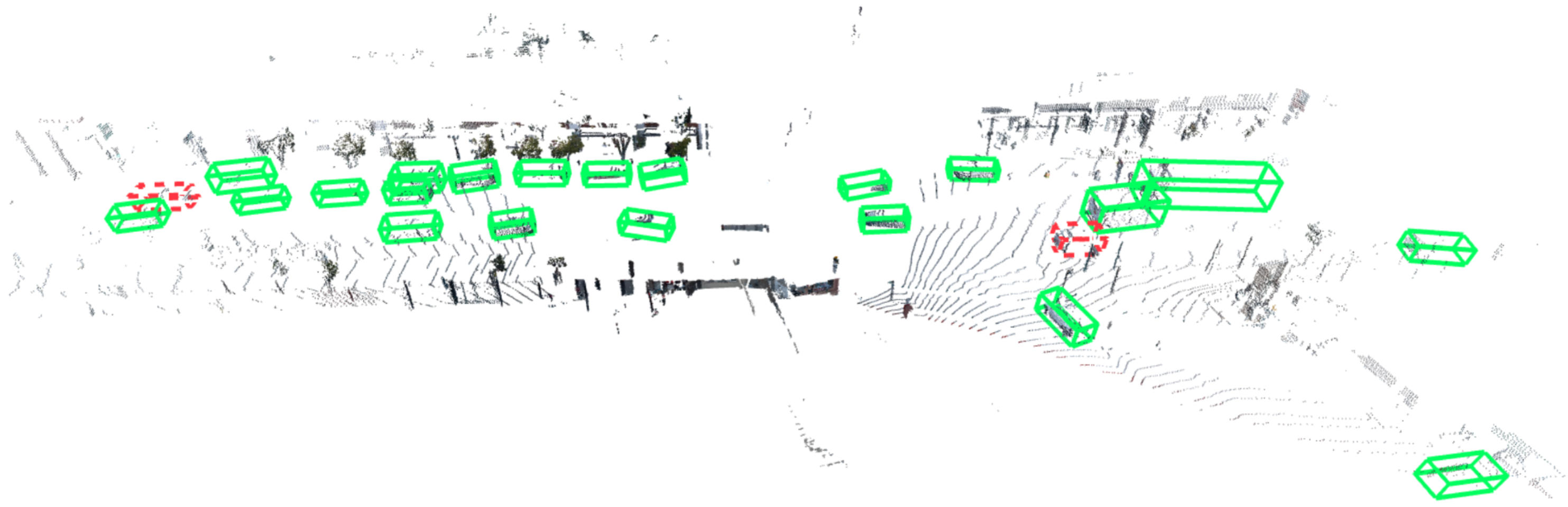}
    \includegraphics[width=0.48\linewidth]{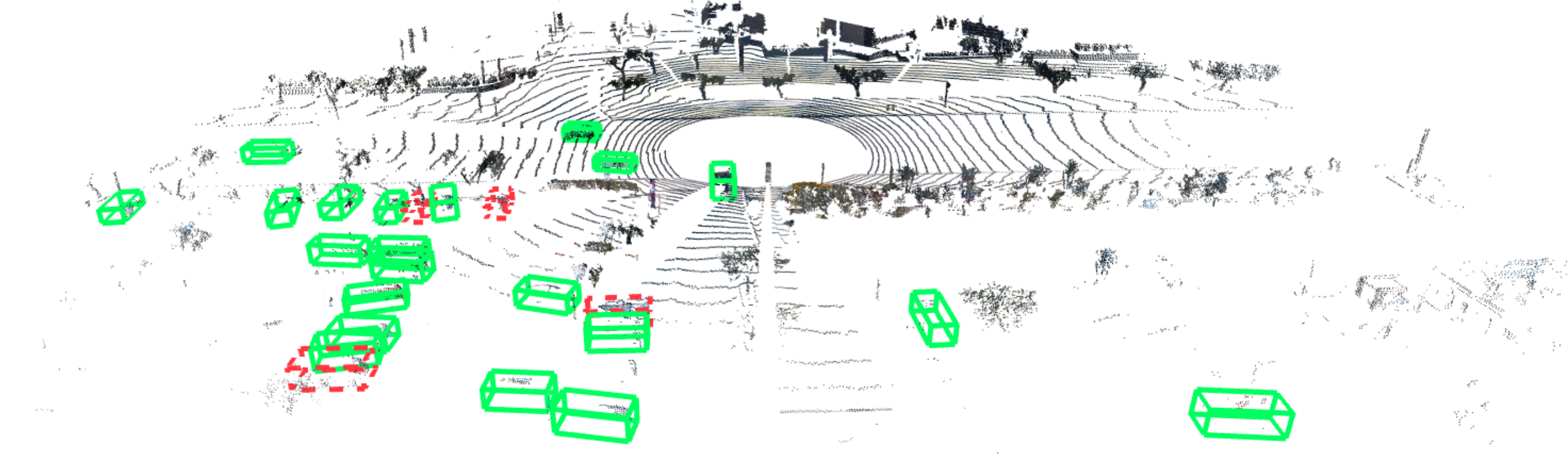}\\
    \includegraphics[width=0.48\linewidth]{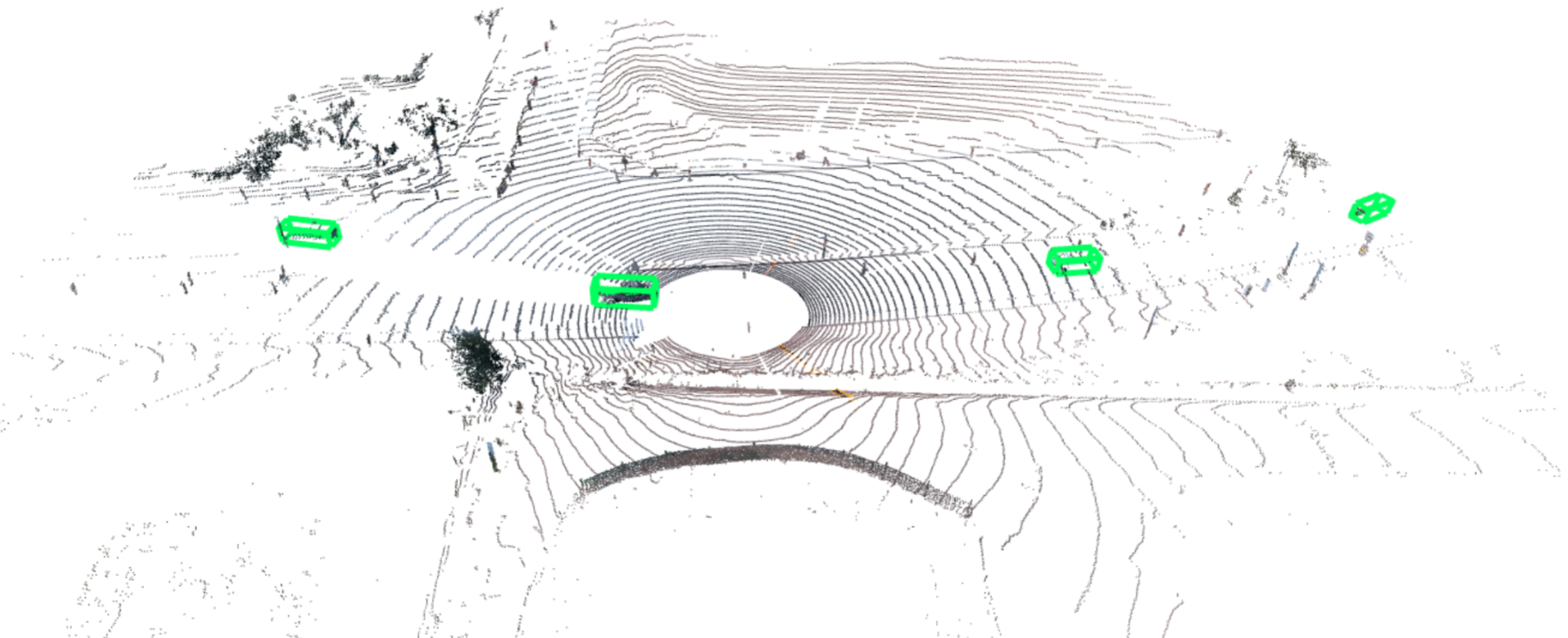}
    \includegraphics[width=0.48\linewidth]{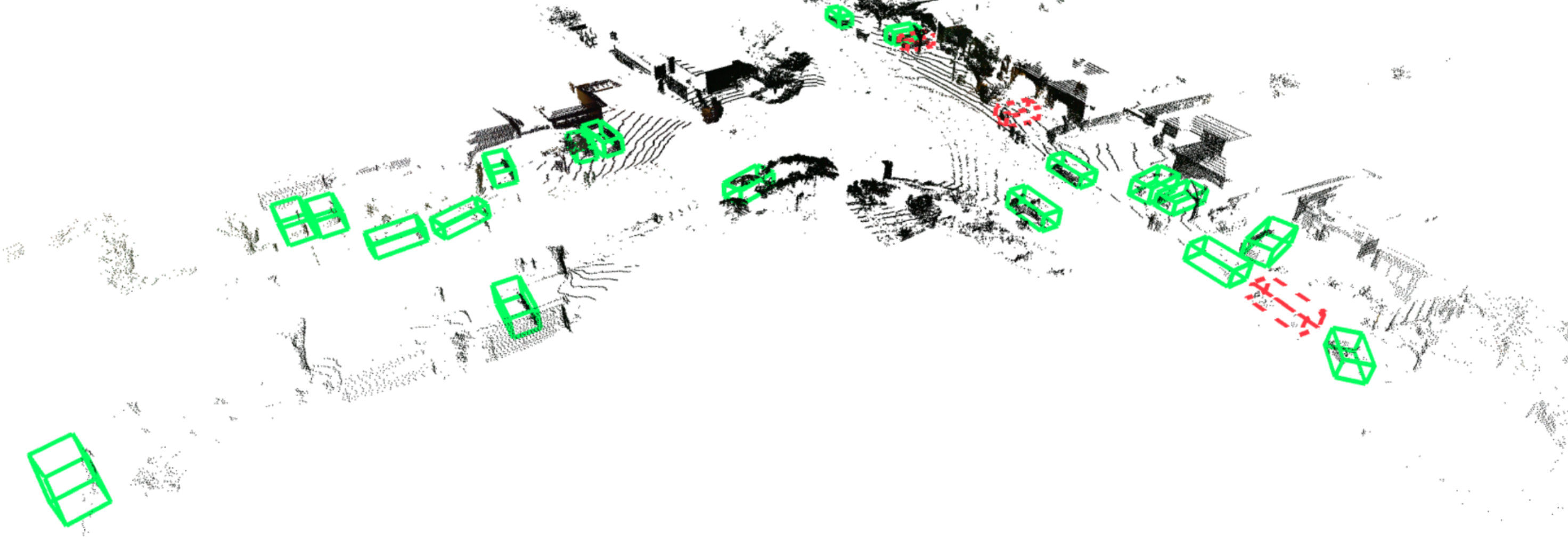}

    \caption{Example results on the Waymo Open Dataset. Green boxes are correct detections, and red boxes indicate errors (dashed lines: FN, solid lines: FP). We observe that the model is more inclined to produce false-negatives, rather than false-positives.}
     \label{fig:waymo_more}
\end{figure*}



\vspace{-0.2cm}
\section{Experiments}
\vspace{-0.1cm}
We conduct experiments on the Waymo Open Dataset ~\cite{sun2019scalability}, following the standard evaluation protocol and evaluation script provided. 
We report results on vehicles at 0.7 IoU overlap, which is the same setting 
as in previous work.
%
We conduct ablation studies which demonstrate the various benefits of the proposed approach,
and provide analysis with respect to different model and input configurations. 
Some of the ablation experiments show additional opportunities for improving performance which are not included in the main results, e.g., increasing the number of RGB frames or input resolutions or using stronger backbones.

\vspace{-0.15cm}
\subsection{Waymo Open Dataset results}
\label{sec:waymo}
\vspace{-0.1cm}
~\autoref{tab:main} shows the results of 4D-Net in comparison to the recent state-of-the-art (SOTA) approaches. As seen, it outperforms the previous SOTA, e.g. by \textbf{3.3 and 4.2 AP}, and more importantly, it significantly outperforms when detecting objects at far distances by \textbf{14.6 and 15.4 AP}. 
We also report inference runtimes, split into time to run the 4D-Net (net) with PC accumulated in time and RGB streams, and the time for point cloud pre-processing (voxelization).
We observe very competitive runtimes, despite processing much more information than other methods.
%
%
Qualitative results are shown in \autoref{fig:WOD} and \autoref{fig:waymo_more}.




\begin{table*}[]
    \centering
     \begin{tabular}{c|ccc|cc|ccccc}
    \toprule
    & \multicolumn{3}{c}{Components} & \multicolumn{2}{c}{Modalities} & \\
    Method & Proj & Conn & Dyn & PC & RGB  & AP L1 & AP L2  & AP 30m & AP 30-50m & AP 50m+ \\
    \midrule
  
    4D-Net & \checkmark & \checkmark & \checkmark & PC+T & RGB+T & \textbf{73.6} & \textbf{70.6} & \textbf{80.7} & \textbf{74.3} & \textbf{56.8} \\
    
     & \checkmark & \checkmark &  & PC+T & RGB+T &73.1 & 70.1 & 80.5  & 73.9 & 56.2 \\
   
     & \checkmark &  &   & PC+T & RGB+T  & 72.6 & 69.7  & 79.6 & 72.8 & 54.6 \\
   
     & & &  & PC+T & RGB+T  & 62.5 & 58.9  & 70.1 & 57.8 & 42.5 \\

    
    \midrule
    
     & \checkmark & \checkmark &  & PC+T  & RGB &72.6 & 69.7 & 79.8 & 73.6 & 55.8 \\

      & \checkmark & \checkmark &  & PC & RGB+T & 65.4 & 63.9 & 77.5  & 67.4 & 48.5 \\
   
     & \checkmark & \checkmark & & PC & RGB &64.3 & 63.0 & 74.9  & 65.1 & 47.2 \\
     
     & \checkmark &  & & PC & RGB & 62.5 & 61.5  & 71.5  & 61.5  & 41.0 \\

     & & \checkmark & & PC & RGB  & 56.7 & 53.6 & 66.2 & 52.5  & 37.5  \\
    
     & & & & PC+T & & 60.5 & 55.7 & 68.4 & 57.6 & 38.1\\
    
     & & & & PC & &55.7 & 52.8 & 65.0 & 51.3 & 35.4 \\

     \bottomrule
    \end{tabular}
       \caption{Ablation results for the 4D-Net. From the full 4D-Net variant, components are removed one at a time, to demonstrate their effect. Proj is the proposed projection method, Conn is the connectivity search and Dyn is the dynamic connection method. We also ablate with using single PC or single RGB input. PC+T denotes Point Clouds in Time, RGB+T is RGB frames in time.
       16 PCiT and 16 RGB of 224x224 size are used (except the top two with 12 RGB of 192x192).
       This is a single 4D-Net. See~\autoref{tab:ablation-multi} for multi-stream 4D-Nets.
       }
    \label{tab:ablation-main}
\end{table*}

\vspace{-0.15cm}
\subsection{Ablation Studies}
\label{sec:ablation}

This section presents the ablation studies.  
We make best efforts to isolate confounding effects and test components individually, e.g., by removing multiple PCs, or RGB. 



    
\begin{table*}[]
    \centering
     \begin{tabular}{l|c|ccc|c}
    \toprule
    Method &AP  &30m &30-50m &50m+ &Runtime \\
    \midrule
    4D-Net (192x192 12f video)  & 73.6  	&80.7	&74.3	&56.8 & 142 ms   \\
    4D-Net MS (192x192 16f video + 312x312 image)  &74.5   &80.9	&74.7	&57.6 & 203 ms  \\
    \midrule
    
    4D-Net MS-1 (192x192 16f video + 224x224 image) &73.4		&81.2	&72.5	&56.5 & 162 ms  \\
    4D-Net MS-2 (224x224 16f video + 224x224 image) &73.8		&80.5	&73.7	&56.9 & 171 ms  \\
    4D-Net MS-3 (128x128 16f video + 192x192 image + 312x312 image) &74.2	&81.5	&72.9	&57.8 & 225 ms \\
    \bottomrule
    \end{tabular}
       \caption{Ablation results for  Multi-Streams (MS) models. AP shown. The top portion shows the main 4D-Net and the Multi-Stream version from \autoref{tab:main}. MS-1 and MS-2 include an additional image stream but at different resolutions. MS-3 has two additional image streams. It shows one can significantly reduce the input video resolution achieving top results. All video models use 16 frames except 4D-Net which has 12. Voxel pre-processing is not included in runtime as in~\autoref{tab:main}.}
    \label{tab:ablation-multi}
\end{table*}

\begin{table}[]
    \centering
    \scalebox{0.917}{
     \begin{tabular}{l|c|ccc|c}
    \toprule
    Image resolutions & AP  & 30m & 30-50m & 50m+ &Time \\
    \midrule
    192x192 1 fr. & 60.8	&73.6	&60.7	&40.4 &82\\
    224x224 1 fr. & 64.3 & 74.9  & 65.1 & 47.2 &97\\
    312x312 1 fr. & 67.3 & 75.7  & 66.4 & 49.5 &142\\
    512x512 1 fr. & 68.2 & 75.9  & 67.5 & 52.4 &297\\
    \midrule
    192x192  12-fr.   &64.2	&75.2	&65.3	&46.2 &109\\
    224x224  16-fr.   & 65.4 & 77.5 & 67.4  & 48.5 &115\\
    \midrule
    224x224  3DRes   & 66.4 & 77.8 & 68.6  & 49.5 &254 \\
    224x224  Assm   &66.8	&79.1 &69.2 &50.7  &502\\
    
    \bottomrule
    \end{tabular}
    }
    \caption{Ablations for input image resolutions for a single frame RGB tower and a single point cloud. AP (L1) shown.
    For comparison, models with 12-frame and 16-frame video input are given, as well as,  stronger but much slower methods 3DResNet~\cite{tran2018closer} and AssembleNet~\cite{ryoo2020assemblenet}, both with 32 frames.
       All are with a single point cloud, which reduces compute, as well. Time is in ms.}
    \label{tab:ablation-resol}
    \vspace{-0.3cm}
\end{table}

\begin{table}[]
    \centering
    \scalebox{0.917}{
     \begin{tabular}{l|c|ccc}
    \toprule
    Number of PC & AP  & AP 30m & AP 30-50m & AP 50m+ \\
    \midrule
    1 PC  & 55.7 & 65.0 & 51.3  & 35.4 \\
    2 PC & 56.3 & 66.1 & 52.5  & 36.4 \\
    4 PC & 57.8 & 66.9 & 54.6  & 36.7 \\
    8 PC & 59.4 & 67.6 & 56.4  & 37.8 \\
    16 PC & 60.5 & 68.4 & 57.6  & 38.1 \\
    32 PC & 60.3 & 67.4 & 56.4  & 38.4 \\
    \bottomrule
    \end{tabular}
    }
       \caption{Ablations for Point Clouds (PC) in time. No RGB inputs are used. One PC (top row) is effectively the PointPillar model.}
    \label{tab:ablation-pc}
    \vspace{-0.cm}
\end{table}

\textbf{Main Contributions and Fusion Approaches.}
~\autoref{tab:ablation-main} shows the main ablation experiments, investigating key components of the 4D-Net. Starting from the main 4D-Net approach (first line), in the top lines we evaluate the approach when the key contributions are removed individually or jointly. As seen, dynamic learning on top of 3D projections is the most beneficial and both are important. At the bottom of the table, we show performance of the approach with different modalities enabled or disabled, for direct comparison. We notice interesting phenomena: using multiple PC in time is definitely helpful, but a single RGB image can boost up the performance of both a single PC and multiple PCs much more significantly, with the proposed projection and connection learning. Similarly, RGB in time can boost a single PC variant significantly, too. Thus a combination of the two sensors, at least one of which is in time is important. The best result comes from multiple RGBs and PC in time, i.e., spanning all 4 dimensions. 
The supp. material has details of the baselines used in lieu of the proposed components.

\textbf{Multi-Stream 4D-Net Variants.}
\autoref{tab:ablation-multi} shows the results of using a various number of additional input streams (\autoref{fig:multi}). As seen, various interesting combinations can be created and learned successfully. Multi-stream models can also afford to reduce the resolution of the video stream inputs and obtain equally powerful models. More multi-stream variants can be explored.

\textbf{RGB Resolution and Video.}
In this section we explore the impact of RGB resolution (\autoref{tab:ablation-resol}\footnote{The AP is lower than the main 4D-Net as we use a single PC and no dynamic connections, to show the effects of image or video models.}). 
We observe potential improvements to the 4D-Net that were not used in the main method and can be leveraged in future work. Some improvements are gained at the expense of runtime as seen to the right of the table.
As expected, better image resolution helps, particularly for detection of far-away objects, e.g., almost 5 and 12 AP improvements for objects beyond 50m when resolution is increased to 512 from 224 and 192. 
Another interesting observation is that RGB video helps a lot. For example, increasing resolution from 224 to 512 improves still-image performance by 1\% for close objects, but keeping the same 224x224 resolution for a video input gets even higher performance, an improvement by 2.6\%.
Naturally, higher resolution videos than the ones shown will improve both metrics at additional latency cost.

Leveraging more powerful video methods is also possible, although not particularly worthwhile, as they gain only a little in accuracy, but at high computational cost (see~\autoref{tab:ablation-resol}, bottom).   
Specifically, we compare two of the most popular video models: the 3D-ResNet~\cite{tran2018closer,hara20173Dresnet} and AssembleNet~\cite{ryoo2020assemblenet}. As seen, they provide more accurate results, especially the powerful AssembleNet, but are very slow. 






\textbf{Point Clouds in Time.}
\autoref{tab:ablation-pc} shows the effect of using point clouds in time. As expected, multiple point clouds improve performance notably. Using 16 PCs, about 1.6 seconds of history, seems to be optimal and also matches observation that density saturates around 16 frames (\autoref{fig:pc-density}).


\vspace{-0.5cm}
\section{Conclusions and Future Work}
\vspace{-0.2cm}
We present 4D-Net, which proposes a new approach to combine underutilized RGB streams with Point-Cloud-in-time information. We demonstrate improved state-of-the-art performance and competitive inference runtimes, despite using 4D sensing and both modalities in time.
%
Without loss of generality, the same approach can be extended to other streams of RGB images, e.g., the side cameras providing critical information for highly occluded objects, or to diverse learnable feature representations for PC or images, or to other sensors.
While this work is demonstrated for the challenging problem of aligning different sensors for autonomous driving which span the 4D, the proposed approach can be used for various related modalities which capture different aspects of the same domain: aligning audio and video data or text and imagery. 


{\small
\bibliographystyle{ieee_fullname}
\bibliography{egbib}
}

\clearpage
\newpage

\section{Implementation Details}
The models were implemented in TensorFlow. We trained for $120\,000$ iterations using a batch size of $256$, split across $8$ devices. The learning rate was set to $0.0015$ using a linear warmup for $6\,000$ steps followed by a cosine decay schedule.

The anchor boxes had size of $[4.7, 2.1, 1.7]$, and $2$ rotations ($0$ and $45$ degrees) used at each feature map location. For the PointPillar pseudo-image creation, we used a grid size of $(224, 224, 1)$, an x-range of $(-74.88, 74.88)$ and the same for y-range. The z-range was $(-5, 5)$. The max number of points per cell, $N=128$. We used 10,000 pillars.

We applied data augmentation to the point clouds (random 3D rotations and flips). The camera matrices were also updated based on the augmentations so the projections would still apply. No augmentation was used on the RGB streams.

The PointPillars network used a feature dim of $64$ for the input points. The created pseudo-image had $224\times 224$ shape. This was followed by $3$ convolutional blocks with $4$, $6$, and $6$ repeats. Each block consisted of a convolution, batch norm and ReLU activation. This was followed by $3$ deconvolutional layers which generate the predictions.

The RGB single frame network is a standard ResNet-18. The video network is based on TinyVideoNetworks. Specifically, we use a network that consists of 6 Residual blocks, the structure is outlined in \autoref{tab:tvn}. Note that the first two blocks apply both spatial and temporal convolutions to the data (in that order).

We trained on the Waymo Open Dataset, which consists of $1\,950$ segments that are each $20$ seconds long (about $200$ frames), a total of $390\,000$ frames. The LiDAR data is processed into point clouds grouped by timestamp and aligned with the RGB frames. In most experiments, we take sequences of $16$ frames as input and predict 3D boxes for the last frame. We evaluate using the provided Waymo metrics library. Before NMS, we filter out boxes with probability less than $0.4$ and boxes larger than $30$m in length and $5$m in width and boxes smaller than $0.5$m in length and width.

\begin{table}[]
    \centering
    \scalebox{0.7}{  
    \begin{tabular}{c|cccc}
    \toprule
     Block & Conv Size & Channels & Repeat & Output Size \\
    \midrule
    Input & - & - & - & 16$\times 224\times 224$ \\
        Block 1 & 1x3x3 + 3x1x1 conv & 32 & 1 & 8$\times 112\times 112$  \\
         Block 2 & 1x3x3 + 3x1x1 conv & 64 & 1 & $4\times 56\times 56 $ \\
         Block 3 & 1x3x3 conv & 128 & 4 & $2\times 28\times 28$  \\
         Block 4 & 1x3x3 conv & 256 & 4 & $2\times 14\times 14$  \\
    \bottomrule
    \end{tabular}
    }
    \caption{RGB Video Network structure used in 4D-Net. Note that the sizes are shown assuming 16 frames at $224\times 224$ input size. For networks that used smaller inputs, the output sizes are each step would be smaller, following the same scaling. Average pooling was used after the convolution to reduce the spatial size. The first two blocks apply both spatial and temporal convolutions in that order.}
    \label{tab:tvn}
\end{table}

\begin{table}[!htbp]
    \centering
    \scalebox{0.7}{  
    \begin{tabular}{l|cc|ccc}
    \toprule
    Method   & AP L1 & AP L2  & AP 30m & AP 30-50m & AP 50m+   \\
    \midrule

Base PointPillars &	55.7 & 52.8 & 65.0 & 51.3 & 35.4 \\
 RGB &	56.7 & 53.6 & 66.2 & 52.5 & 37.5 \\
 Spatial Avg & 58.9 & 57.6 & 69.7 & 56.1 & 39.6 \\
 Spatial Transformer & 59.5 & 58.2 & 70.0 & 54.7 & 43.1\\ 
 Projection &	64.3 & 63.0 & 74.9 & 65.1 & 47.2\\
 \bottomrule
 \end{tabular}
 }
    \caption{Comparison of different spatial fusion methods.}
    \label{tab:fuse}
\end{table}

\section{Additional experimental results}
\paragraph{Fusion Methods} In the main paper, we focused on projection as the main method to fuse RGB and point cloud data. Here, we also compare to several other methods. The results are shown in \autoref{tab:fuse}.

\emph{Basic RGB.} Here we flatten the RGB image feature output into a 1-D tensor, and concatenate it to the point cloud feature. This loses all spatial information and essentially puts the entire image into each point.

\emph{Spatial Avg Pooling.} As another baseline, we apply average spatial pooling to the image-based features $R_i$, obtaining a $F^R_i$-dimensional feature vector. We then concatenate this to each feature in the PC-based features $M_i$, resulting in a $(X^M_i, Z^M_i, F^M_i+F^R_i)$ feature map. This is then passed through the remaining CNN for classification. This provides the point cloud stream with some RGB information, but it has no spatial information.

\emph{Spatial Transformer.} We also tried using a spatial transformer \cite{jaderberg2015spatial} to crop regions around each projected point to append to the PC feature. However, despite small improvements, we found this to be extremely slow due to taking many spatial crops with the transformer.

These baselines are compared with the proposed \emph{Projection} method which is in Table 2 of the main paper. The experiments are conducted in the same conditions as the Projection method in the main paper. The \emph{Basic PointPillars} (also in Table 2 of the main paper) does not have an RGB input and is included for reference only.

\section{Additional Visualizations}
\begin{figure*}
\vspace{-0.1cm}
 
     \includegraphics[width=1.0\linewidth]{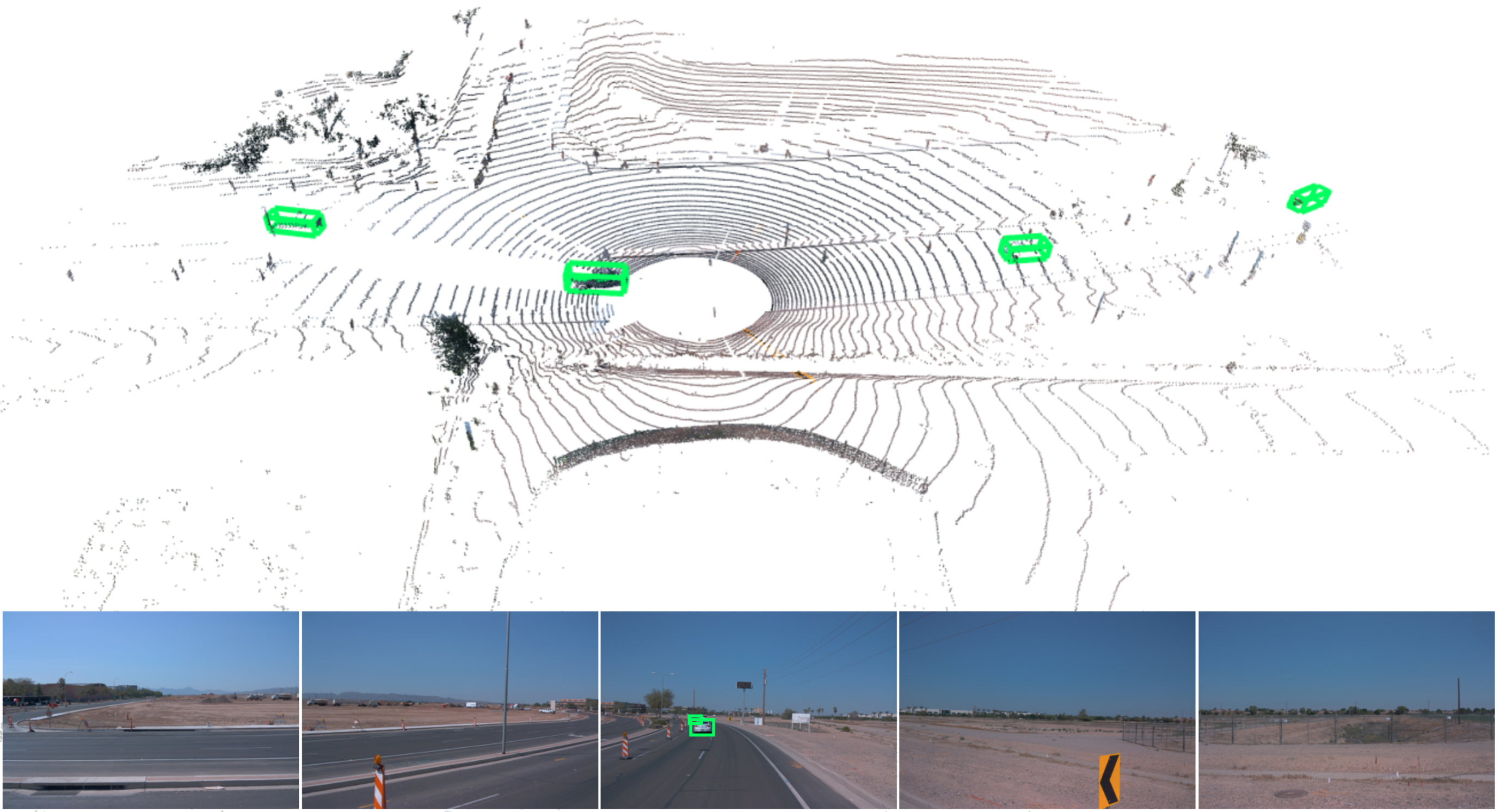}
    \includegraphics[width=1.0\linewidth]{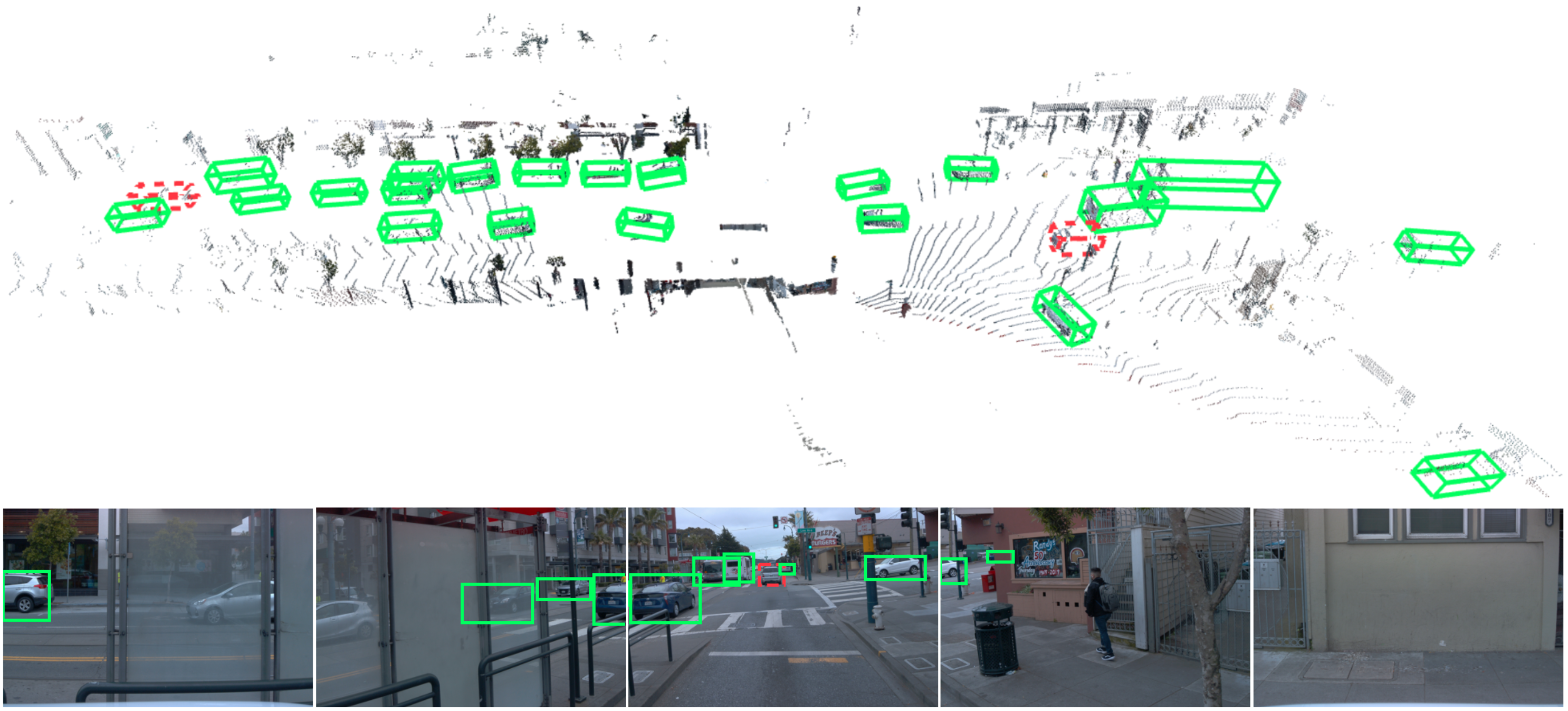}   
    \caption{4D-Net predictions on a scene in the Waymo Open Dataset~\cite{sun2020scalability}. Individual instances are shown in green (here) or in different colors in the figures below. Red boxes indicate errors (dashed lines: FN, solid lines: FP). 
    The front camera (central image) is the only one used in our work presently, the others are included for visualization purposes.
    Note that any misalignments in the camera view are due to projection, not by inaccuracies in the predictions.}
     \label{fig:WODSUPP}
\vspace{-3mm}
\end{figure*}

In \autoref{fig:WODSUPP} we show more visualizations of the predictions of the 4D-Net on the Waymo Open Dataset.

\begin{figure*}
    \centering

    \includegraphics[width=1.0\linewidth]{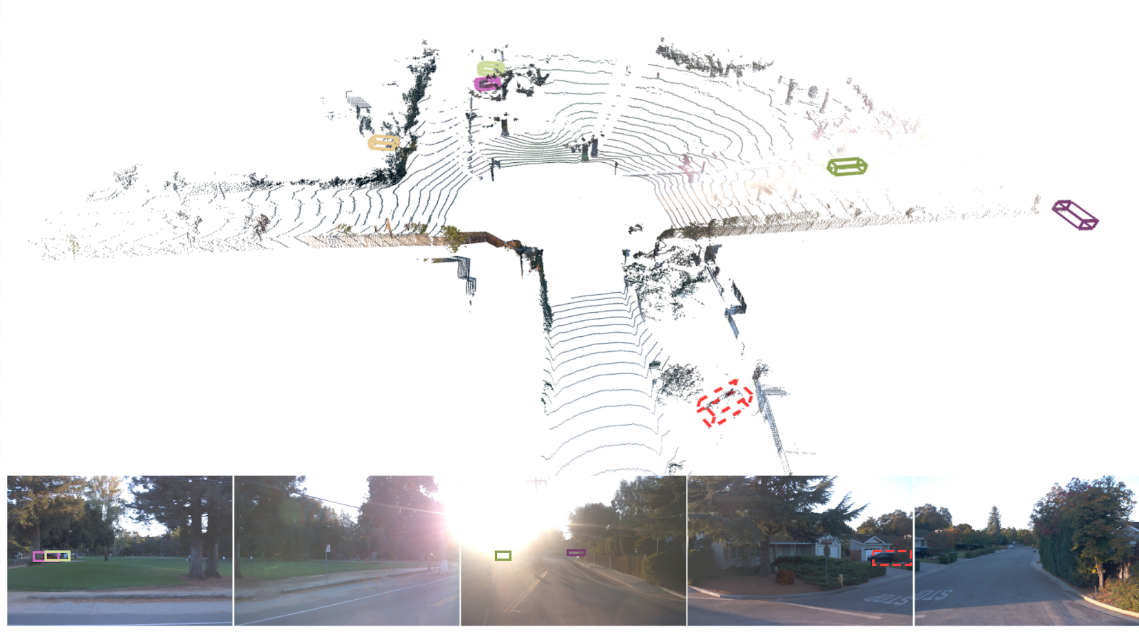}

    \includegraphics[width=1.0\linewidth]{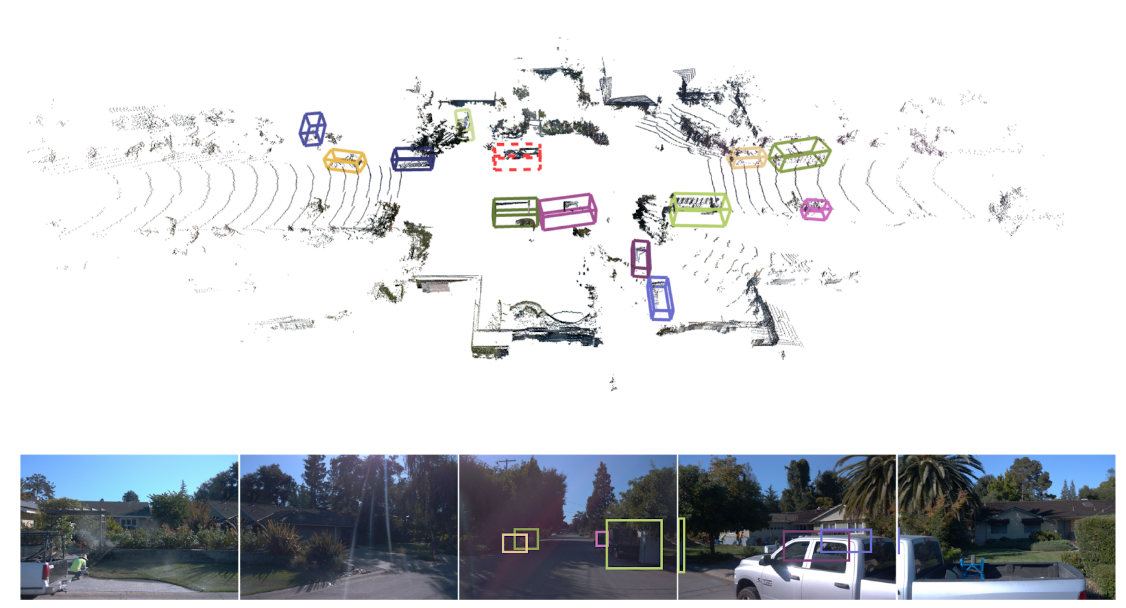}
\end{figure*}

\begin{figure*}
   \centering

    \includegraphics[width=1.0\linewidth]{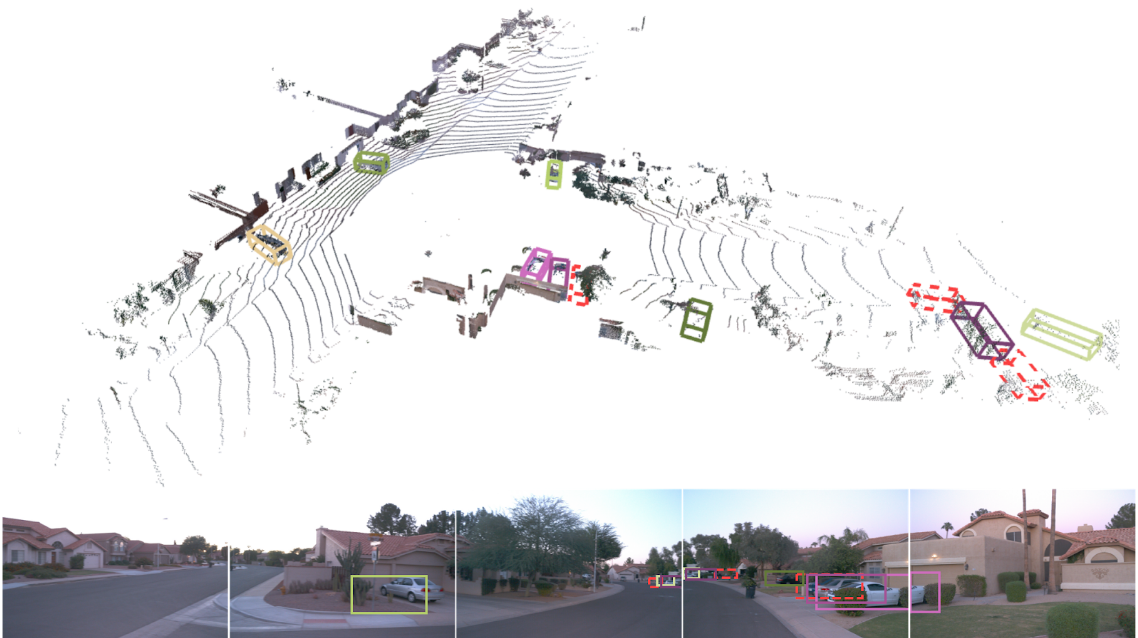}
    \includegraphics[width=1.0\linewidth]{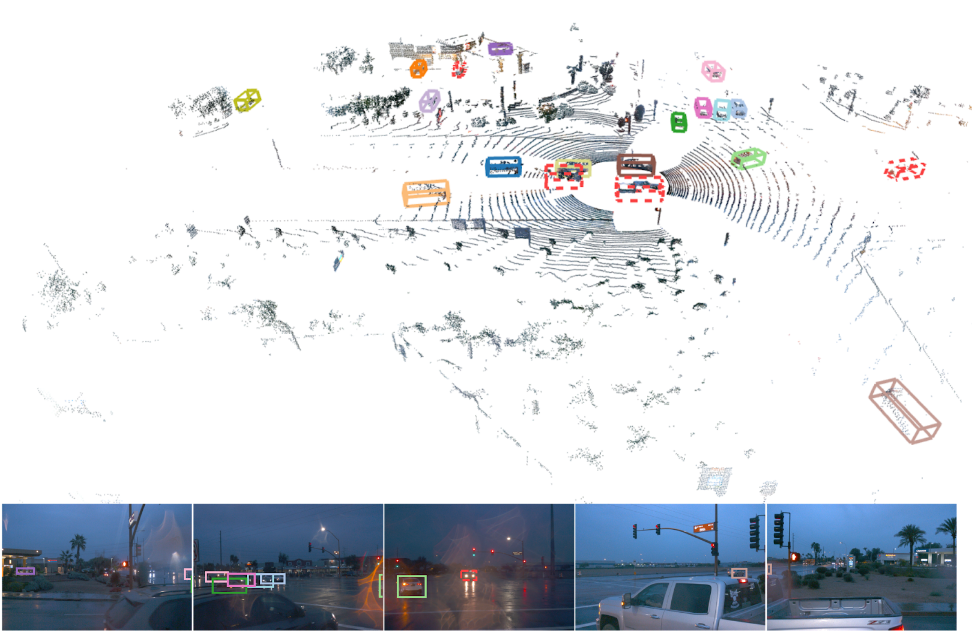}
\caption{Failure cases examples: In these two challenging cases, the central image (stream) is not sufficient and a vehicle is missed.}
\end{figure*}

\end{document}